\documentclass[10pt,twocolumn,letterpaper]{article}

\usepackage{iccv}
\usepackage{times}
\usepackage{epsfig}
\usepackage{graphicx}
\usepackage{amsmath}
\usepackage{amssymb}

\usepackage{subfigure}
\usepackage{tabularx}
\usepackage{makecell}

\usepackage[pagebackref=true,breaklinks=true,letterpaper=true,colorlinks,bookmarks=false]{hyperref}

\iccvfinalcopy 

\def\iccvPaperID{2} 

\ificcvfinal\fi

\begin{document}

\title{CNN-based Cost Volume Analysis as Confidence Measure for Dense Matching}

\author{Max Mehltretter \qquad Christian Heipke\\
Leibniz University Hannover, Germany\\
{\tt\small \{mehltretter,heipke\}@ipi.uni-hannover.de}
}

\maketitle
\ificcvfinal\thispagestyle{empty}\fi

\begin{abstract}
Due to its capability to identify erroneous disparity assignments in dense stereo matching, confidence estimation is beneficial for a wide range of applications, e.g. autonomous driving, which needs a high degree of confidence as mandatory prerequisite.
Especially, the introduction of deep learning based methods resulted in an increasing popularity of this field in recent years, caused by a significantly improved accuracy. Despite this remarkable development, most of these methods rely on features learned from disparity maps only, not taking into account the corresponding 3-dimensional cost volumes. However, it was already demonstrated that with conventional methods based on hand-crafted features this additional information can be used to further increase the accuracy.
In order to combine the advantages of deep learning and cost volume based features, in this paper, we propose a novel Convolutional Neural Network (CNN) architecture to directly learn features for confidence estimation from volumetric 3D data.
An extensive evaluation on three datasets using three common dense stereo matching techniques demonstrates the generality and state-of-the-art accuracy of the proposed method.
\end{abstract}

\section{Introduction}

\begin{figure}
\centering

\subfigure[]
{
    \includegraphics[width=0.75\linewidth]{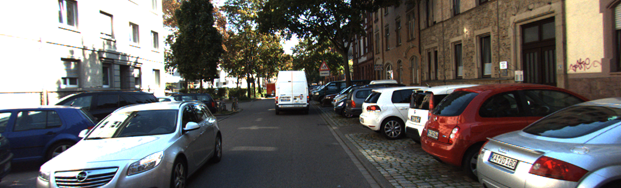}
}
\subfigure[]
{
    \includegraphics[width=0.75\linewidth]{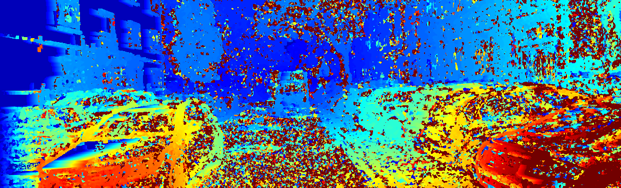}
}
\subfigure[]
{
    \includegraphics[width=0.75\linewidth]{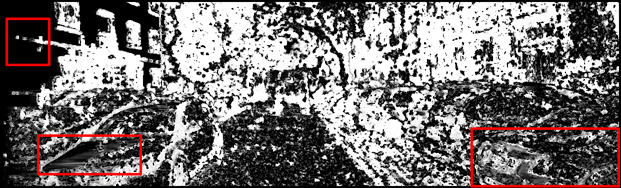}
}
\subfigure[]
{
    \includegraphics[width=0.75\linewidth]{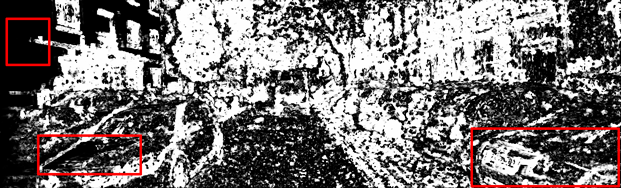}
}
\caption{\textbf{Confidence estimation on an image of KITTI 2015 dataset} \cite{Menze2015}. (a) Reference image, (b) corresponding disparity map computed with Census \cite{Zabih1994}, (c) confidence map computed with a CNN learned on disparity maps and (d) with the proposed method. The red boxes highlight noisy and low-texture regions where the proposed method outperforms the state-of-the-art.}
\label{fig:IntroExample}
\end{figure}

The reconstruction of depth information from a stereoscopic image pair is a classical task in 
photogrammetry and computer vision and the minimal case of the well-known structure from motion problem. A special case of this task is dense image matching. It not only determines depth for significant feature points, but for every or at least a large majority of pixels within an image pair. In principle, depth reconstruction can be interpreted as inverse operation to a perspective projection, which directly leads to the major difficulty of this task: Projecting the 3D scene to a 2D image plane results in a dimensionality reduction. Consequently, the inverse operation does not have a unique solution in general, characterising it as ill-posed. To determine a solution nevertheless, the identification of point correspondences within the two images of a pair is a prerequisite in general.
This raises the question about the reliability of such a solution. Especially under challenging conditions, depth reconstruction approaches might not be able to identify the correct correspondences for all pixels. Thus, it is particularly important to be able to identify cases in which high uncertainty exists regarding the result. This task is referred to as confidence estimation and has been the subject of many investigations in recent times.

All of the proposed methods have in common that they estimate the confidence pixel by pixel. This allows to filter out local outliers from disparity maps and thus to subsequently adjust the ratio of density and reliability.
The proposed applications are diverse: Confidence maps are used as weighting-schemes to combine multiple stereo matching algorithms \cite{Poggi2016,Spyropoulos2015}, different cost functions \cite{Batsos2018,Hu2012,Mehltretter2018a} or to fuse cost volumes \cite{Sun2017} for multi-view stereo, in a reasonable way.
Confidence maps are furthermore used to improve the process of depth reconstruction itself: They allow to modulate cost functions in order to adjust the influence of a specific disparity assignment on its neighbours during optimisation \cite{Park2015,Spyropoulos2014}. Finally, also Semiglobal Matching (SGM) \cite{Hirschmuller2008} can be improved by introducing a confidence-based adjustment of the penalties \cite{Sun2017} or by applying a weighted aggregation scheme for combining different paths within the optimisation process \cite{Poggi2016a}.

Together with those applications, a variety of different approaches were proposed to estimate confidence. In this context, the introduction of deep learning based procedures can be seen as a milestone, improving the accuracy significantly. But at the same time, the diversity of the utilised information has decreased. While many hand-crafted features were designed based on a variety of different 2D and 3D cues, nowadays features are mainly learned on disparity maps and corresponding reference images only.
Nevertheless, the usage of hand-crafted features has already proven that especially the information contained in 3-dimensional cost volumes resulting from the cost computation step of dense image matching can be beneficial.
These cost volumes provide the cost distribution over the whole disparity range instead of just the supposedly optimal value. 

Consequently, in this paper we present an approach which operates directly on those 3-dimensional cost volumes. Contrary to the assumptions made in \cite{Kim2019} and \cite{Seki2016}, we demonstrate the advantages of using raw cost volumes as input to a confidence estimation network (cf. Fig.\,\ref{fig:IntroExample}).
Thus, the main contributions of this work are:
\begin{itemize}
	\item A CNN-based approach for estimating the confidence of a disparity assignment based on the cost curves of a pixel and its neighbours. For this purpose, an architecture is presented which allows to learn features directly from the volumetric 3D cost volume.
	\item An extension to the commonly used Area Under the Curve (AUC) evaluation approach, which minimises the error introduced by discretisation.
	\item An extensive evaluation showing the accuracy and generality of our approach. For this purpose, the performance is examined on three well-established datasets regarding three popular stereo matching methods and compared against the state-of-the-art.
\end{itemize}
\section{Related Work}

\subsection{Confidence Estimation}\label{sec:rw-ce}
With growing popularity in recent years, the approaches to estimate confidence of disparity assignments became as diverse as their applications, but can in principle be divided into 3 groups:
The first group is based on individual, hand-crafted features. For this purpose, e.g. the properties of a cost curve, the consistency between disparity estimations in the left and right image and the distinctiveness of a pixel in its local neighbourhood are used.
It is noteworthy that in particular methods which are based on the characteristics of cost curves show convincing results. This statement is supported by the high accuracy of a recently published approach, where ambiguous solutions are penalised based on the position and distinctiveness of the global minimum compared to local ones \cite{Veld2018}. A good overview of the commonly used features is given in \cite{Hu2012}.

The approaches of the second group combine certain of those features to form more accurate and robust measures. Beside linear aggregation \cite{Sun2017}, random forest based combinations are especially popular \cite{Batsos2018, Haeusler2013, Park2015, Poggi2016a, Spyropoulos2015}. The transition to the third group is accomplished by utilising neuronal networks to carry out the combination task \cite{Poggi2017b, Seki2016}.

Finally, the approaches within the third group map the whole confidence estimation process to convolutional neural networks. For this purpose, \cite{Poggi2016b} (CCNN) as well as \cite{Poggi2016, Poggi2017} utilise square patches extracted from disparity maps and centred on a pixel of interest to determine its confidence. \cite{Seki2016} in addition, proposes to stack two of those patches, one from the left, one from the right image, in order to introduce the idea of left-right-consistency. On the other hand, \cite{Fu2017} suggests to combine patches from disparity maps and the RGB reference image to increase the available amount of information (LFN).
Lately, \cite{Tosi2018} presented an approach (LGC-Net) utilising not only information from a local neighbourhood, but also from global context. For this purpose, a two-part network architecture is proposed, which uses a local component \cite{Fu2017, Poggi2016b} to detect high frequency changes and an encoder-decoder based module to enlarge the receptive field.

Analysing the advantages of the different approaches, two main points are noticeable: On the one hand, features from cost volumes demonstrate superior performance. On the other hand, learned features outperform hand-crafted ones. \cite{Kim2019} and \cite{Shaked2017} combine those assumptions and take a first step towards learning features on cost volumes. While they propose different approaches, both contain a preprocessing step to extract subsets of data, which are provided as input to their networks. They state that such a preprocessing is necessary since the cost distributions of raw cost volumes do not allow to distinguish between correct and incorrect estimations in general. However, the proposed preprocessing steps limit the information provided to the confidence estimation step. This prevents the method from exploiting the full potential of learning features on cost volumes.

Since the methods CCNN, LFN and LGC-Net, discussed in this section, represent the state-of-the-art by estimating confidence with the highest accuracy, in Section\,\ref{sec:eval} they are used for comparison with our method.

\subsection{Deep Learning on Volumetric Data}
Confidence estimation based on a cost volume can be interpreted as a regression task on volumetric 3D data, since it is the prediction of real numbers within the unit interval.
In the literature, mainly two types of methods exist for processing volumetric 3D data: projection-based approaches \cite{Johns2016, Shi2015, Su2015} and voxel-based processing \cite{Maturana2015, Wu2015}.
The former is based on the idea to project 3D data to one or multiple 2D images and apply classification in 2D, using well established network architectures. Benefiting from the extensive research on 2D image classification, those methods demonstrated superior performance compared to voxel-based approaches for many applications. However, they mainly classify samples based on an object shape and surface. For the present task of evaluating cost volumes, this is not reasonable, since these volumes always have the same shape and only the values within the individual voxels vary.

The performance gap to voxel-based methods was mainly caused by the higher complexity and increased memory consumption of learning features from 3D directly.
In recent years, multiple approaches were published to overcome this gap. In \cite{Qi2016} a hybrid method is proposed, combining projection- and voxel-based ideas: On the one hand, auxiliary losses are introduced by additionally classifying subvolumes. On the other hand, the 3D volume is reduced to a 2D image by applying convolutional layers similar to X-ray scanning, subsequently allowing the application of conventional image-based CNNs. On the contrary, the methods in \cite{Li2016a} and \cite{Riegler2017} benefit from the sparsity of 3D data transferred to a volumetric representation.

However, those approaches are not applicable to cost volumes, since they assume a completely different type of data. Cost volumes are dense voxel grids and in general, their subvolumes have limited expressiveness. Along their depth axis, they consist of a single cost curve, potentially leading to different results if only sections are examined.
\section{CNN-based Cost Volume Analysis}

\begin{figure}
\centering

\subfigure[]
{
    \includegraphics[width=0.45\linewidth]{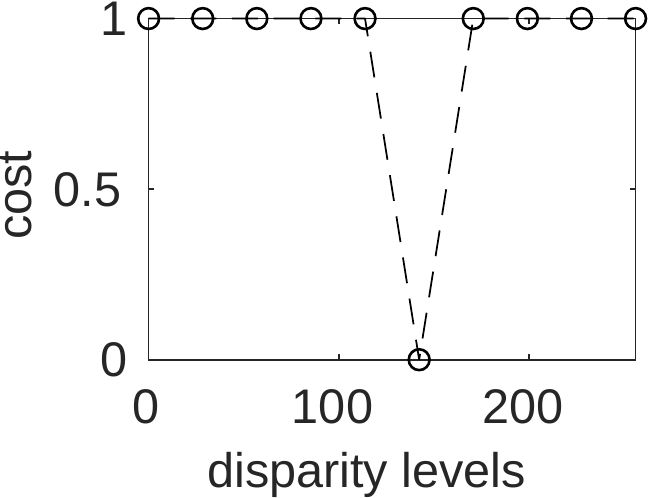}
    \label{fig:CostCurves-Optimal}
}\hfill
\subfigure[]
{
    \includegraphics[width=0.45\linewidth]{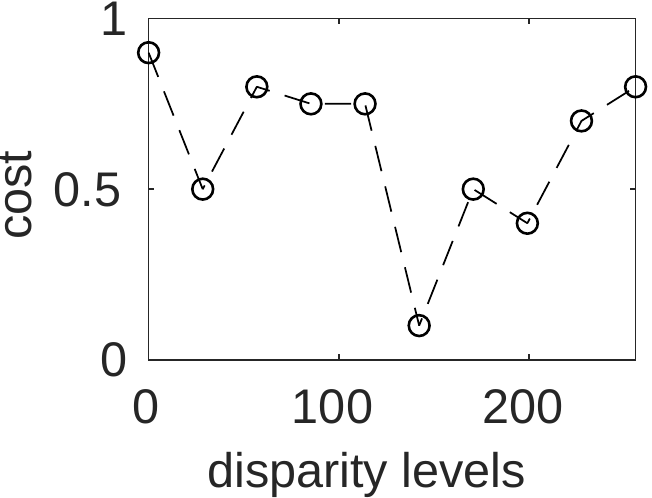}
    \label{fig:CostCurves-ClearMin}
}
\subfigure[]
{
    \includegraphics[width=0.45\linewidth]{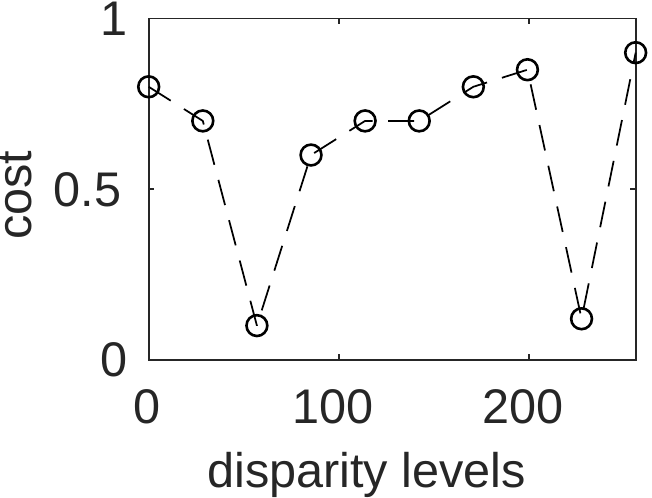}
    \label{fig:CostCurves-2Min}
}\hfill
\subfigure[]
{
    \includegraphics[width=0.45\linewidth]{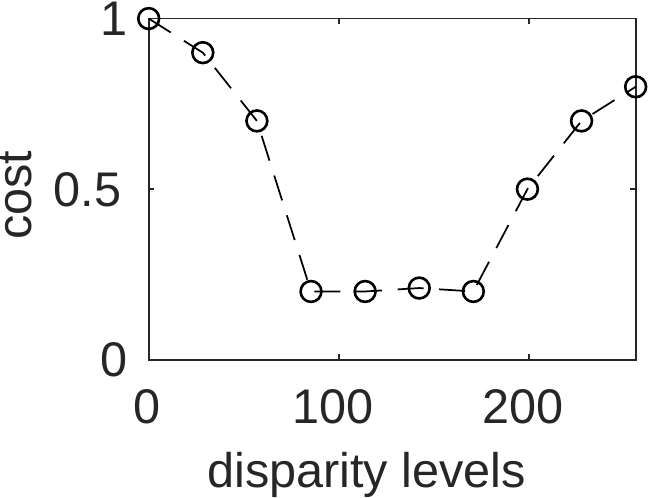}
    \label{fig:CostCurves-SaddlePoint}
}
\caption{\textbf{Exemplary cost curves demonstrating the relation between cost function and confidence.} (a) An ideal curve, characterised by a single minimum with zero cost and all other values being one. (b) A more realistic curve with multiple minima, but a reliably identifiable global minimum, still results in high confidence. (c) No distinct global minimum is identifiable, making the determination of the correct correspondence unreliable - a typical behaviour in areas with repetitive patterns. (d) The occurrence of a wide and flat minimum is a typical behaviour in non-textured areas and leads to an inaccurate localisation of the correct correspondence. The latter two cases result in low confidence.}
\label{fig:CostCurves}
\end{figure}

The main idea of the approach presented in this work is to assess the confidence of a disparity map pixel by pixel based on the corresponding cost volume. In this context, a cost volume is the result of the cost computation step of an arbitrary stereo matching approach carried out on an epipolar rectified image pair. 
The axes $x$ and $y$ of such a cost volume correspond to the image coordinates, while the disparity axis $z$ represents the associated cost curves.

In general, typical characteristics can be observed on cost curves, independent from their source: In the ideal case, the cost curve contains a unique minimum with zero cost, while all other values are at maximum (Fig.\,\ref{fig:CostCurves-Optimal}). In practice, however, cost curves usually have several local minima, requiring the theoretical assumptions to be relaxed. A disparity assignment with high confidence is characterised by a clearly identifiable and unambiguous global minimum (Fig.\,\ref{fig:CostCurves-ClearMin}). In contrast, low confidence is usually assigned if either no distinct global minimum can be identified (Fig.\,\ref{fig:CostCurves-2Min}) or if the global minimum is wide and flat, making the localisation of the correct correspondence inaccurate (Fig.\,\ref{fig:CostCurves-SaddlePoint}).

\subsection{Cost Volume Normalisation}\label{sec:norm}

As stated in \cite{Kim2019}, cost curves highly depend on the utilised stereo matching approach.
Consequently, a unified data representation is a prerequisite for learning to estimate confidence directly from 3D cost volumes.
For this purpose, the theoretical boundaries of the matching approach's result space are used to nlormalise the cost volumes. The result is a 3D tensor of real values in the range [0,1].

\begin{figure*}
    \centering
    \includegraphics[width=\linewidth]{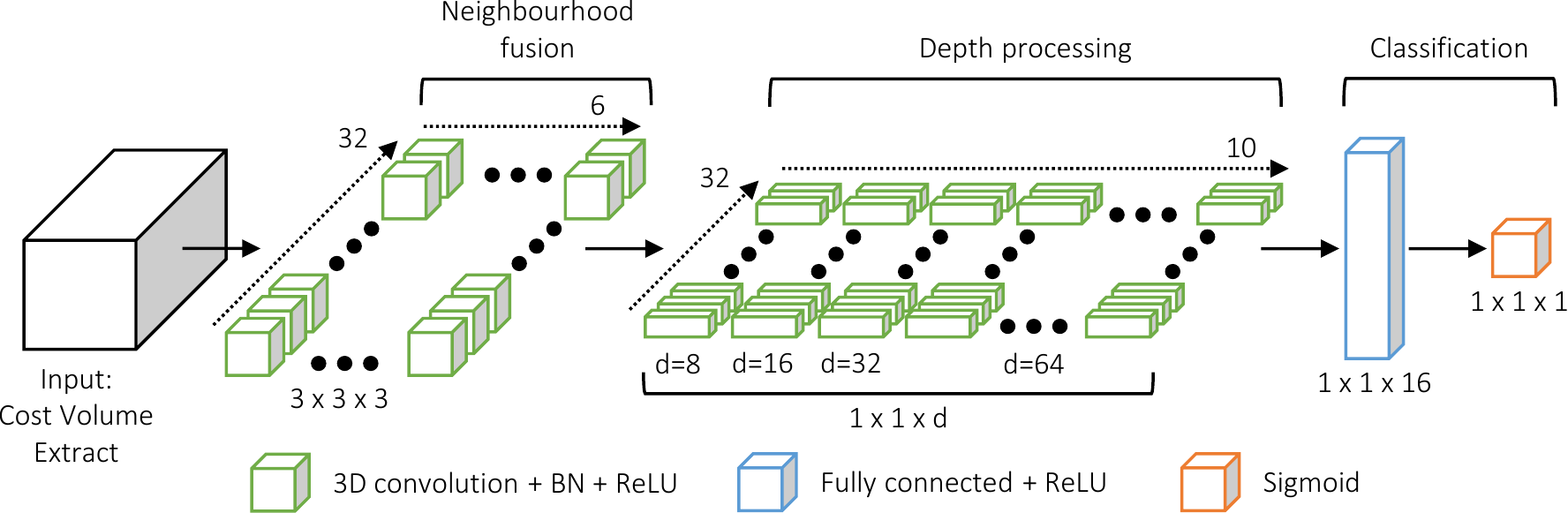}
    \caption{\textbf{Our confidence estimation network, CVA-Net} (\underline{C}ost \underline{V}olume \underline{A}nalysis \underline{Net}work).
    Consisting of three main elements, the network first fuses a cost volume extract into a single cost curve. This curve is then processed along the disparity axis by convolutions with varying depth. The fully-connected layers at the end of the network perform the classification by estimating the confidence.}
\label{fig:architecture}
\end{figure*}

\subsection{Architecture}

To address the task of confidence estimation from normalised 3D cost volumes, we introduce a novel CNN architecture referred to as \textit{Cost Volume Analysis Network} (CVA-Net).
It follows the idea of a feedforward network and consists of three main components: neighbourhood fusion, depth processing and classification (c.f. Fig.\,\ref{fig:architecture}). A detailed layer-by-layer definition can be found in Tab.\,\ref{table:architecture}. 
As input, the network takes cost volume extracts of size ${N\times N\times D}$. The size of the perceptive field is set to $N=13$\,pixels, providing a good trade-off between the amount of information available to the network and the degree of smoothing within the resulting confidence map.
In order to contain complete cost curves, the depth $D$ of an extract is chosen to be equal to the depth of the cost volume. In the specific case, the depth of a volume is set to $256$\,pixels according to the theoretical maximum disparity of the training samples.

The first part of the network, the neighbourhood fusion, merges the information contained in an extract to a single cost curve. The basic idea behind this procedure is equivalent to that of most region-based matching approaches: Including neighbourhood information increases the robustness. Especially if the cost curve corresponding to the pixel of interest is affected by noise or delivers an ambiguous solution, neighbourhood information is beneficial. The depth of the filters associated to this part of the network is set to $3$ to handle minor shifts of the curves, e.g. caused by discretisation errors during the cost computation step.

\begin{table}
	\centering
	\footnotesize
	\caption{\textbf{Summary of the proposed CVA-Net architecture.} Unless otherwise specified, each layer is followed by batch normalisation (BN) and a ReLU non-linearity.}
	\vspace{3mm}
	\begin{tabularx}{\linewidth}{l|X|c}
        \ Layer & Description & \makecell{Output Tensor\\ Dimensions}\\
        \hline\hline
        \ Input & Cost Volume Extract & 13$\times$13$\times$256\\
        \hline
        \multicolumn{3}{c}{\textbf{Neighbourhood Fusion}}\\
        \hline
        \ 1 & 3D conv., 3$\times$3$\times$3, 32 filters & 11$\times$11$\times$254\\
        \ 2 & 3D conv., 3$\times$3$\times$3, 32 filters & 9$\times$9$\times$252\\
        \ 3 & 3D conv., 3$\times$3$\times$3, 32 filters & 7$\times$7$\times$250\\
        \ 4 & 3D conv., 3$\times$3$\times$3, 32 filters & 5$\times$5$\times$248\\
        \ 5 & 3D conv., 3$\times$3$\times$3, 32 filters & 3$\times$3$\times$246\\
        \ 6 & 3D conv., 3$\times$3$\times$3, 32 filters & 1$\times$1$\times$244\\
        \hline
        \multicolumn{3}{c}{\textbf{Depth Processing}}\\
        \hline
        \ 7 & 3D conv., 1$\times$1$\times$8, 32 filters, zero padding & 1$\times$1$\times$244\\
        \ 8 & 3D conv., 1$\times$1$\times$16, 32 filters, zero pad. & 1$\times$1$\times$244\\
        \ 9 & 3D conv., 1$\times$1$\times$32, 32 filters, zero pad. & 1$\times$1$\times$244\\
        \ 10-16 & 3D conv., 1$\times$1$\times$64, 32 filters, zero pad. & 1$\times$1$\times$244\\
        \hline
        \multicolumn{3}{c}{\textbf{Classification}}\\
        \hline
        \ 17 & Fully-connected, 16 nodes, no BN & 1$\times$1$\times$16\\
        \ 18 & Fully-connected, 1 node, sigmoid non-linearity, no BN & 1$\times$1$\times$1\\
  \end{tabularx}
	\label{table:architecture}
\end{table}

In the subsequent depth processing part, the merged cost curve is further processed in order to derive high-level features characterising the curve.
It is noteworthy that the filter depth $d$ increases with the layer depth: Starting with $d=8$ the value is doubled with every new layer until $d=64$ is reached. Our experiments have shown that this design performs slightly better than one with a constant filter depth while having to learn a significantly smaller number of parameters. Furthermore, zero padding is utilised for all convolutions in the depth processing part of the network. This keeps the size of the output tensor constant and, compared to no padding, provides a greater number of features as input for the subsequent confidence estimation.

The third and last part of the network consists of fully-connected layers and performs the final confidence estimation. For this purpose, a binary classification of the disparity estimation into \textit{correct} and \textit{incorrect} is carried out, based on the features originating form the depth processing part. The result of the sigmoid non-linearity used for this classification is finally interpreted as confidence and assigned to the centre pixel of the initial cost volume extract.

Following a common procedure \cite{Poggi2016b, Zbontar2016}, the fully-connected layers within the last part are replaced by convolutional layers, transforming the proposed architecture to a fully convolutional network. This allows to train on image patches while computing a confidence map of the full resolution image within a single forward pass during test time. However, this of course also allows for piecewise processing the cost volume if hardware restrictions have to be taken into account.
Similar to the observations described in \cite{Maturana2015}, we found that a network with a single fully-connected layer consisting of a small number of nodes is sufficient for the task of classifying volumetric 3D data. This is certainly also supported by the fact that only binary classification takes place. Since the size and number of convolutional layers is decisive for the quality of the results, these contain the majority of the 782,725 parameters.

\subsection{Training Procedure}\label{sec:training}

Following the training protocol proposed in \cite{Poggi2017a}, we train our network on the first 20 training image pairs of the KITTI 2012 dataset \cite{Geiger2012}. 
For this purpose, tensors of size ${13\times 13\times 256}$ are extracted from normalised cost volumes (see Section\,\ref{sec:norm}) corresponding to the left image of each pair. Every extract is centred on a pixel with available ground truth disparity, resulting in more than 2.7 million training samples. 
Experiments with varying numbers of training samples have shown a convergence of the test accuracy of our network at around 2.6 million samples when trained on the KITTI 2012 dataset. Since the point of convergence strongly depends on the variance of characteristics present in the training samples, this number can vary greatly if the network is trained on different data.

Knowing only whether a disparity estimate is correct or not, the task of estimating its confidence is commonly transferred to a binary classification task. While the network classifies the disparity estimation as \textit{correct} or \textit{incorrect}, the result of the final sigmoid non-linearity is used as confidence score. The ground truth for this binary label is derived from the error metric proposed in \cite{Menze2015}: A disparity estimation $d_{est}$ is assumed to be correct if either $|d_{est}-d_{gt}|<3$\,pixels or $|d_{est}-d_{gt}|<(d_{gt}\times 0.05)$, where $d_{gt}$ is the corresponding ground truth disparity.

Our network is trained on batches of size $256$ for 10 epochs with a learning rate of $10^{-4}$, followed by 3 epochs with a learning rate decreased by factor 10.
While the convolutional layers are initialised with a normal distribution $\mathcal{N}(0,0.0025)$, for the fully-connected layers Glorot initialisation \cite{Glorot2010} is used. 
Adam \cite{Kingma2014} is employed to minimise the Binary Cross Entropy, setting the moment estimates exponential decay rates to their default values $\beta_1=0.9$ and $\beta_2=0.999$. Finally, to enforce generalisation, dropout \cite{Srivastava2014} is applied to the fully-connected layers with a rate of $0.5$.
\section{Experimental Results}\label{sec:eval}
In this section, an extensive evaluation is presented, in which we validate our approach on the datasets KITTI 2012 \cite{Geiger2012}, KITTI 2015 \cite{Menze2015} and Middlebury v3 \cite{Scharstein2014}.
Moreover, the evaluation is carried out on the cost volumes computed by the popular stereo matching methods Census-based block matching (with a support region size of ${5 \times 5}$) \cite{Zabih1994}, Census-based SGM \cite{Hirschmuller2008} and MC-CNN fast \cite{Zbontar2016}. By validating on a local, a global and a deep learning based stereo matching method computed on different datasets, the general validity of the proposed CVA-Net is investigated.
The results are compared against the state-of-the-art confidence estimation methods CCNN \cite{Poggi2016b}, LFN \cite{Fu2017} and LGC-Net \cite{Tosi2018}, already introduced in Section\,\ref{sec:rw-ce}.
To allow a fair comparison, all examined methods have been trained on the same data, following the procedure described in Section\,\ref{sec:training}.

Based on the proposed network design, the depth of a cost volume to be processed is limited to a disparity range of [0,255] pixels. This range was chosen according to the disparities that may occur in the KITTI dataset, which was used to train the proposed CVA-Net.
As a consequence, also the cost volumes processed during inference must conform to this depth resolution. For this purpose, within this evaluation the resolution of the reference images is halved as long as the maximum disparity exceeds 255. Alternatively, a compression of the cost volume would also be conceivable. However, this is beyond the scope of the present work and is the subject of further investigations.

\subsection{Evaluation Protocol}\label{sec:eval-prot}

\begin{figure*}
    \centering
    \includegraphics[width=0.93\linewidth]{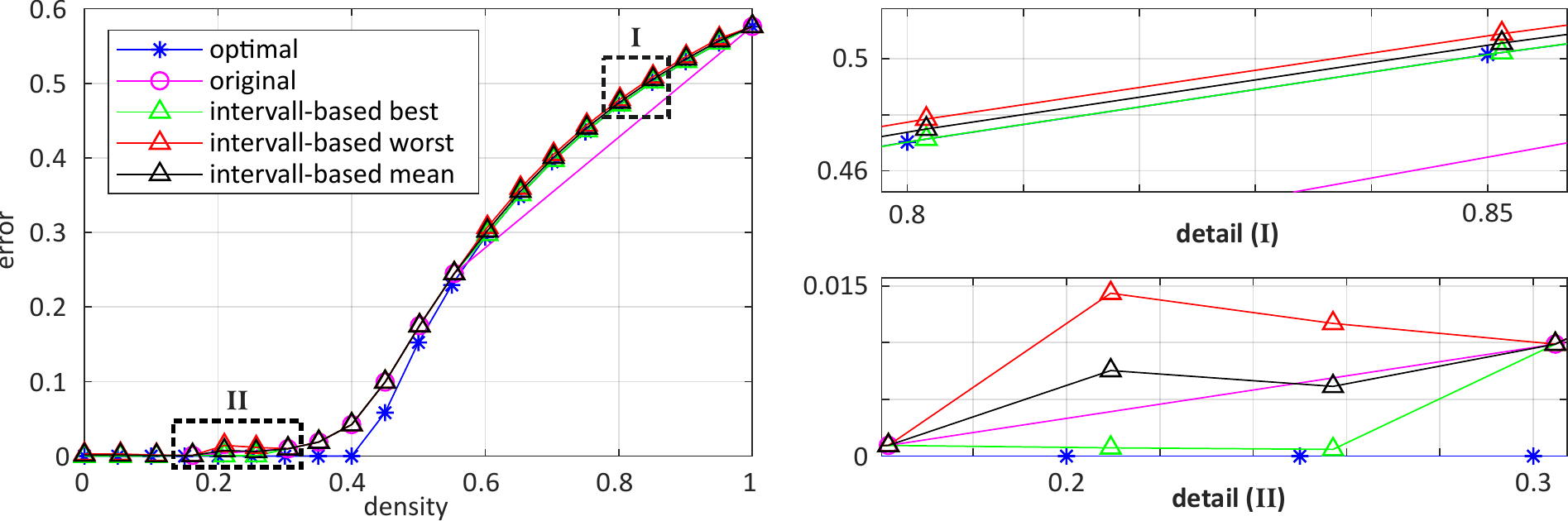}
    \caption{\textbf{Visual proof of the necessity for a refined evaluation method.} 
    Following the original procedure, the ROC curve is not sampled within ranges of the same confidence score. This may result in cases with high discretisation error and segments below the theoretically optimal curve (here especially visible in the upper right part). To minimise the discretisation error, additional sampling points are introduced within those ranges. Ambiguities are avoided by estimating the error of such an additional point based on its lower and upper error bound.}
    \label{fig:opt-auc}
\end{figure*}

To evaluate the methodology presented in this paper, the performance is assessed using a measure which relies on ROC curve analysis. This is a well-established procedure in the field of confidence estimation, originally proposed in \cite{Hu2012}.
The ROC curve represents the error rate as a function of the percentage of pixels sampled from a disparity map in order of decreasing confidence. 
More precisely, the density is sampled in 5\,\% steps, but pixels with equal confidence are processed together to avoid ambiguous results. However, this procedure can lead to large distances in between two sample points (more than 40\,\% of all pixels, c.f. upper right part in Fig.\,\ref{fig:opt-auc}) if many pixels share the same confidence. The consequence is a high discretisation error. In extreme cases, the sampled ROC curve is below the theoretically optimal curve - a contradiction in terms.

As a solution, we propose an interval-based extension to sample additional points within this regions, again in 5\,\% steps. The interval boundaries are defined by the best and worst possible case (all correct/incorrect depth estimates first) and are determined for every additional point. The error of a point is then taken to be the centre of the corresponding interval.
In the last step, the Area Under the Curve (AUC) is computed, which is then used to assess the accuracy of a confidence estimation regarding the detection of wrong disparity assignments. 
Assuming that an optimal confidence map contains higher values for every correct disparity assignment than for any incorrect one, the optimal AUC depends only on the overall error $\epsilon$ of a disparity map:
\begin{equation}
\label{eq:AUC_opt}
	\begin{aligned}
		AUC_{opt} = \int_{1-\epsilon}^{1}\frac{p-(1-\epsilon)}{p}dp\\=\epsilon+(1-\epsilon)\ln(1-\epsilon)
	\end{aligned}
\end{equation}
where $p$ is the percentage of pixels sampled from a disparity map. The closer the AUC of a confidence map reaches the optimal value, the higher the accuracy.

\subsection{Validation on KITTI 2012 \& 2015}\label{sec:eval-kit}

Following the evaluation procedure of recently published confidence measures \cite{Fu2017,Poggi2016b,Tosi2018}, we first assess the performance of the proposed CVA-Net on the KITTI 2012 \cite{Geiger2012} and KITTI 2015 \cite{Menze2015} stereo datasets. Both were captured using vehicle mounted stereo camera set-ups and provide LIDAR based ground truth disparity maps with disparities for 30\,\% of the pixels.
Containing various street scenes from urban as well as rural environments, these datasets still pose a challenge to dense stereo matching algorithms.
Since we perform training using KITTI 2012 images, the 23 images used for training and validating the networks are excluded from the evaluation of the KITTI 2012 dataset, resulting in 171 images for the latter task. From the KITTI 2015 dataset all 200 images are used for the evaluation.

\begin{table}
	\centering
	\footnotesize
	\caption{\textbf{Quantitative results on the three evaluated datasets.}
	The single entries show the theoretically optimal (\textit{Opt.}) and the average AUC $\times 10^2$ of the evaluated confidence measures on the three examined stereo matching methods over all considered images of a dataset. The smaller the values, the better, while \textit{Opt.} is the best achievable value (c.f. Sec.\,\ref{sec:eval-prot}).}
	\vspace{3mm}
	\begin{tabular}{l|c|ccc|c}
		\hline
        \ \makecell{\textit{avg. AUC}\\$=10^{-2}\times$} & Opt. & \makecell{CCNN\\\cite{Poggi2016b}} & \makecell{LFN\\\cite{Fu2017}} & \makecell{LGC-Net\\\cite{Tosi2018}} & \makecell{Ours}\\\hline\hline
        \multicolumn{6}{c}{\textbf{KITTI 2012} \cite{Geiger2012}}\\\hline
        \ CENSUS-BM & 10.94 & 12.37 & 12.30 & 11.97 & \textbf{11.52} \\
        \ CENSUS-SGM & 0.92 & 2.41 & 2.44 & 2.31 & \textbf{2.25} \\
        \ MC-CNN & 2.24 & 2.89 & 2.91 & 2.71 & \textbf{2.55} \\\hline
        \multicolumn{6}{c}{\textbf{KITTI 2015} \cite{Menze2015}}\\\hline
        \ CENSUS-BM & 9.07 & 10.59 & 10.49 & 10.18 & \textbf{9.86} \\
        \ CENSUS-SGM & 0.84 & 2.36 & 2.40 & 2.39 & \textbf{2.31} \\
        \ MC-CNN & 2.46 & 3.35 & 3.35 & 3.19 & \textbf{3.02} \\\hline
        \multicolumn{6}{c}{\textbf{Middlebury v3} \cite{Scharstein2014}}\\\hline
        \ CENSUS-BM & 6.69 & 9.01 & 9.12 & 8.36 & \textbf{8.21} \\
        \ CENSUS-SGM & 2.26 & 5.58 & 6.14 & \textbf{5.33} & 5.40 \\
        \ MC-CNN & 3.54 & 5.22 & 5.27 & 4.91 & \textbf{4.85} \\
		\hline
  \end{tabular}
	\label{table:results}
\end{table}

\begin{figure*}
    \centering
    \subfigure[Reference image]
    {
        \includegraphics[width=0.31\linewidth]{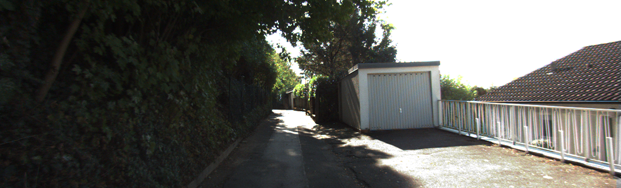}
        \label{fig:QualRes-RefImg}
    }\hfill
    \subfigure[Disparity map (Census-BM)]
    {
        \includegraphics[width=0.31\linewidth]{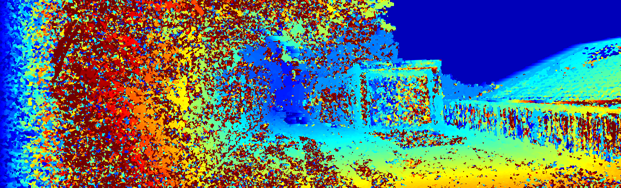}
        \label{fig:QualRes-Disp}
    }\hfill
    \subfigure[Confidence map (CCNN \cite{Poggi2016b})]
    {
        \includegraphics[width=0.31\linewidth]{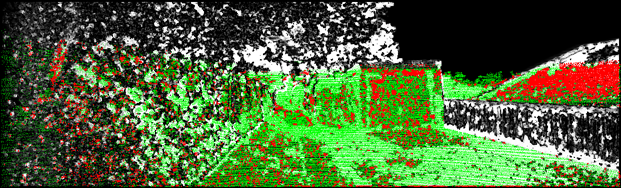}
        \label{fig:QualRes-CCNN}
    }
    \subfigure[Confidence map (LFN \cite{Fu2017})]
    {
        \includegraphics[width=0.31\linewidth]{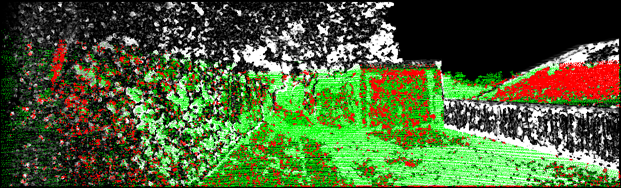}
        \label{fig:QualRes-LFN}
    }\hfill
    \subfigure[Confidence map (LGC-Net \cite{Tosi2018})]
    {
        \includegraphics[width=0.31\linewidth]{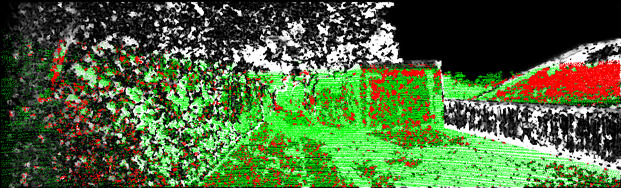}
        \label{fig:QualRes-LGC}
    }\hfill
    \subfigure[Confidence map (ours)]
    {
        \includegraphics[width=0.31\linewidth]{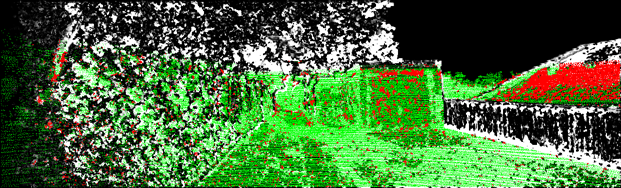}
        \label{fig:QualRes-CVA}
    }
	\caption{\textbf{Qualitative evaluation on frame 90 of the KITTI 2012 dataset} \cite{Geiger2012}.
	The colour of a pixel represents its confidence from black (low) to white (high).
	Using a threshold of $\tau=0.5$, pixels with available ground truth disparity are coloured green if either the assigned disparity is correct and the confidence $c$ is larger than $\tau$ or if the disparity assignment is wrong and $c\leq\tau$. Red pixels, on the other hand, indicate an incorrect confidence estimation.
	The advantages of the proposed CVA-Net can especially be seen in noisy areas of the disparity map, e.g. on the street in the central lower part of the image and in the left part of the image, showing vegetation.}
	\label{fig:EvalKITTI12}
\end{figure*}

\begin{figure*}
    \centering
    \subfigure[Reference image]
    {
        \includegraphics[width=0.31\linewidth]{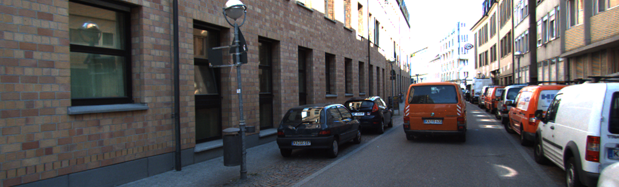}
    }\hfill
    \subfigure[Disparity map (Census-BM)]
    {
        \includegraphics[width=0.31\linewidth]{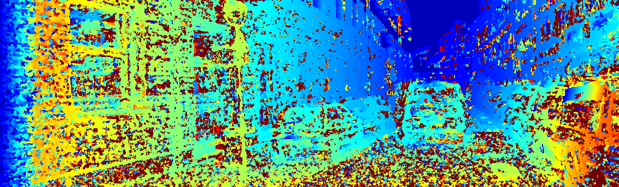}
    }\hfill
    \subfigure[Confidence map (CCNN \cite{Poggi2016b})]
    {
        \includegraphics[width=0.31\linewidth]{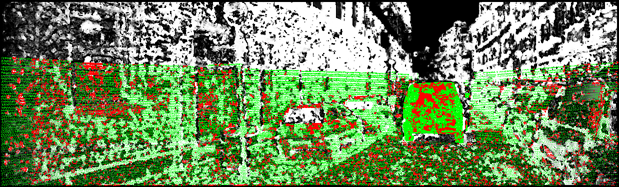}
    }
    \subfigure[Confidence map (LFN \cite{Fu2017})]
    {
        \includegraphics[width=0.31\linewidth]{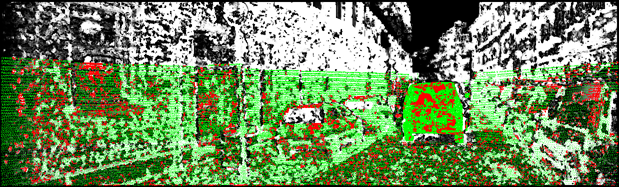}
    }\hfill
    \subfigure[Confidence map (LGC-Net \cite{Tosi2018})]
    {
        \includegraphics[width=0.31\linewidth]{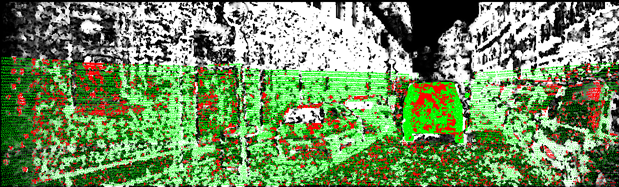}
    }\hfill
    \subfigure[Confidence map (ours)]
    {
        \includegraphics[width=0.31\linewidth]{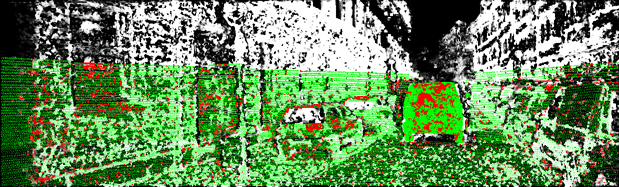}
    }
	\caption{\textbf{Qualitative evaluation on frame 165 of the KITTI 2015 dataset} \cite{Menze2015}.
	For details on the colour coding, please refer to Fig.\,\ref{fig:EvalKITTI12}.
	CVA-Net shows a superior accuracy in noisy regions of the disparity map, especially noticeable for street and car at the right image border. Both regions are characterised by low texture in the reference image, identifiable using the information of the cost curve (c.f. Fig.\,\ref{fig:CostCurves-SaddlePoint}).}
	\label{fig:EvalKITTI15}
\end{figure*}

Analysing the results presented in Tab.\,\ref{table:results}, it can be seen that the proposed approach outperforms the other methods on almost all evaluated configurations.
Especially in noisy regions of a disparity map, superior accuracy can be observed (c.f. Fig.\,\ref{fig:EvalKITTI12} and \ref{fig:EvalKITTI15}). 
This applies regardless of whether these noisy disparity estimates belong to strongly textured (e.g. the vegetation on the right in Fig.\,\ref{fig:QualRes-RefImg}) or low-texture areas (e.g. the street) of the reference image.
Keeping in mind, that all other methods estimate the confidence based on the disparity map (and the reference image for LFN) only, it is evident that our method benefits from the additional information contained in cost volumes along the disparity axis. This statement is also supported by the fact that LGC-Net uses a much wider receptive field ($48\times48$) compared to our method ($13\times13$) and is therefore provided with more information along the height and width axes, but nevertheless performs slightly worse.

On the other hand, Tab.\,\ref{table:results} also illustrates that although the cost volumes are normalised the performance improvements vary between the different stereo matching methods, which is particularly evident in the case of SGM.
This is a clear indication that CVA-Net is sensitive to differences in the characteristics of cost curves resulting from different stereo matching methods.
Nevertheless, the results still proof that the proposed method is applicable to cost volumes of quite different stereo matching approaches.

\subsection{Cross-validation on Middlebury v3}\label{sec:eval-mid}

\begin{figure*}
    \centering
    \subfigure[Reference image]
    {
        \includegraphics[width=0.31\linewidth]{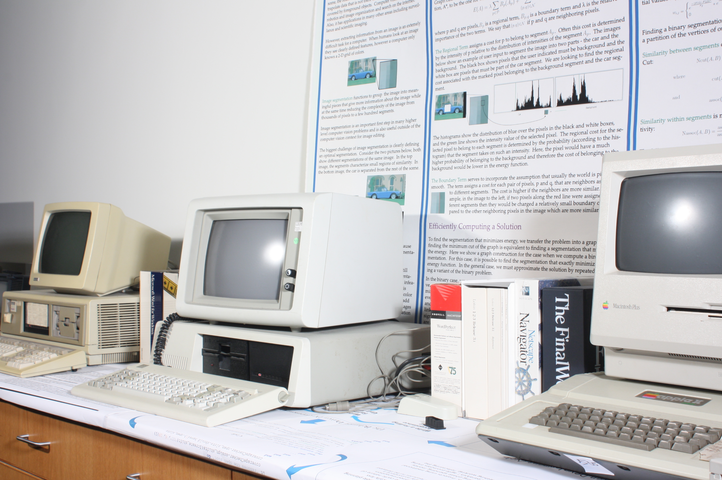}
        \label{fig:QualRes-Mv3-RefImg}
    }\hfill
    \subfigure[Disparity map (Census-BM)]
    {
        \includegraphics[width=0.31\linewidth]{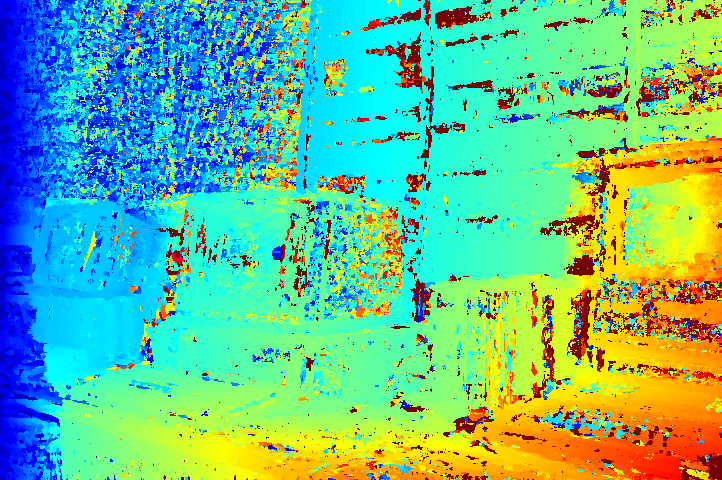}
    }\hfill
    \subfigure[Confidence map (CCNN \cite{Poggi2016b})]
    {
        \includegraphics[width=0.31\linewidth]{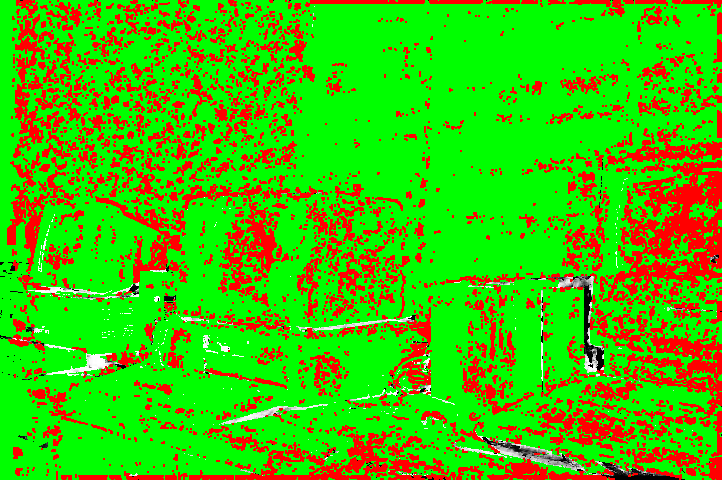}
        \label{fig:QualRes-Mv3-CCNN}
    }
    \subfigure[Confidence map (LFN \cite{Fu2017})]
    {
        \includegraphics[width=0.31\linewidth]{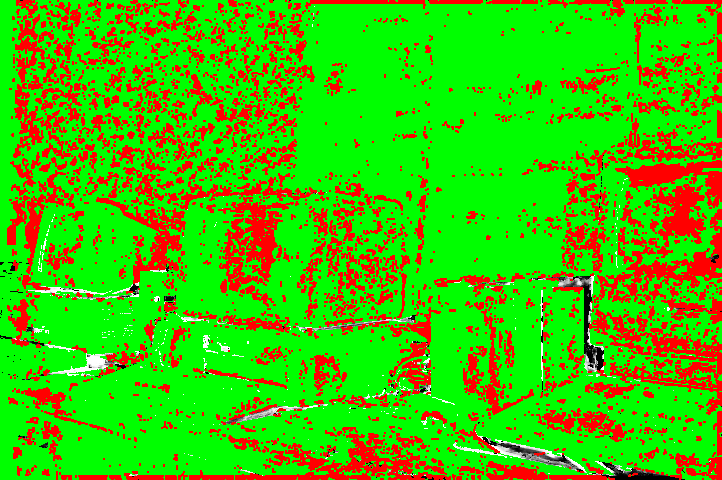}
        \label{fig:QualRes-Mv3-LFN}
    }\hfill
    \subfigure[Confidence map (LGC-Net \cite{Tosi2018})]
    {
        \includegraphics[width=0.31\linewidth]{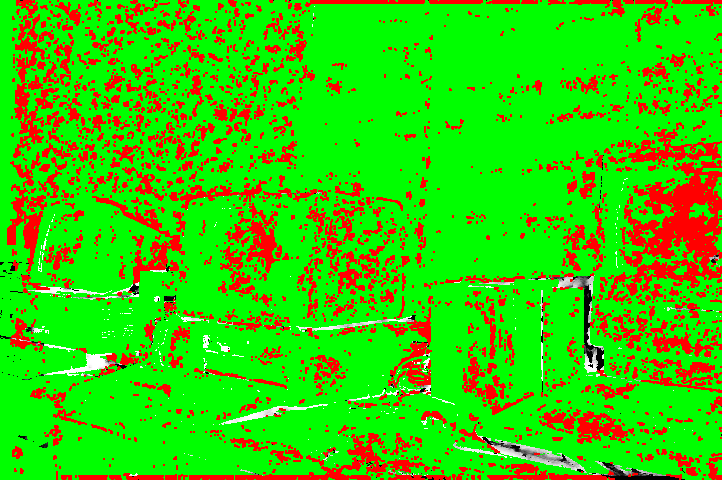}
        \label{fig:QualRes-Mv3-LGC}
    }\hfill
    \subfigure[Confidence map (ours)]
    {
        \includegraphics[width=0.31\linewidth]{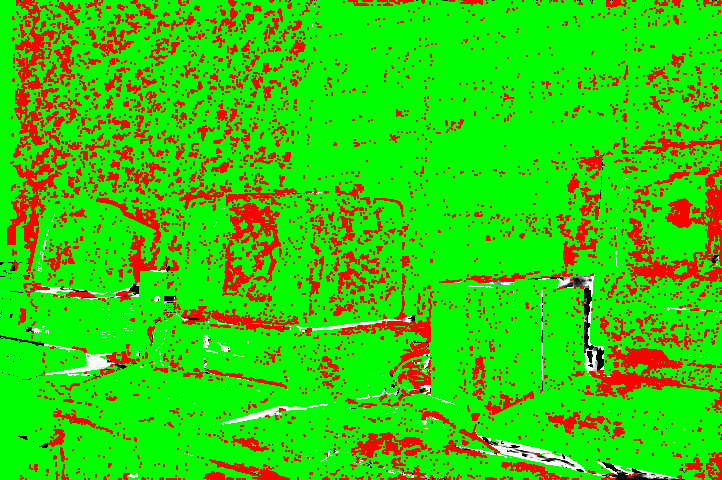}
        \label{fig:QualRes-Mv3-CVA}
    }
	\caption{\textbf{Qualitative evaluation on frame 'Vintage' of the Middlebury v3 dataset} \cite{Scharstein2014}.
	For details on the colour coding, please refer to Fig.\,\ref{fig:EvalKITTI12}.
	This example proofs visually that CVA-Net's ability to estimate the confidence of a disparity assignment generalises well over different datasets. 
	It can be observed that the proposed network can also detect the ambiguity of texture-less regions if they are represented relatively smoothly and with low noise within the disparity map (e.g. on the right screen). 
	On the other hand, it can also be seen that most of CVA-Net's erroneous confidence estimates are close to depth discontinuities, indicating one of its limitations.}
	\label{fig:EvalMv3}
\end{figure*}

After demonstrating the performance of our network on the dataset it was trained on, we now illustrate the generality of our solution by testing it on images showing completely different environments. For this purpose, we evaluate the same three stereo matching methods on the Middlebury v3 dataset \cite{Scharstein2014}. It contains 15 training samples showing various indoor scenes captured with a static stereo set-up and providing dense ground truth disparity maps based on structured light. Due to the limitation of the cost volume depth explained at the beginning of this section, the images of the Middlebury dataset are processed at one quarter of the original resolution.
Contrary to the KITTI benchmark, the Middlebury benchmark accepts disparity assignments $d_{est}$ to be correct if 
$|d_{est}-d_{gt}|\leq1$\,pixel, where $d_{gt}$ is the corresponding ground truth disparity.
However, since the error metric specified with the KITTI datasets is utilised for training, to ensure consistency, it is also used for evaluating the confidence estimations on the Middlebury v3 dataset.

Similar to the results on the KITTI datasets, the proposed CVA-Net shows state-of-the-art accuracy on the Middlebury dataset as well (c.f. Tab.\,\ref{table:results}). This proofs that the concept of learning to estimate the confidence of a disparity assignment based on its cost curve generalises well over different datasets.
As illustrated in Figure\,\ref{fig:EvalMv3}, also on the images of the Middlesbury dataset CVA-Net is characterised by the ability to estimate confidence accurately in noisy areas. The following example furthermore demonstrates the advantage of using the information of the cost curve: The texture-less region around the specular reflection on the right computer screen in Fig.\,\ref{fig:QualRes-Mv3-RefImg} corresponds to a relatively smooth area within the disparity map. If only the disparity map is used, it is highly challenging to assess the confidence for this case correctly (c.f. Fig.\,\ref{fig:QualRes-Mv3-CCNN} and \ref{fig:QualRes-Mv3-LGC}). As can be seen in Fig.\,\ref{fig:QualRes-Mv3-LFN}, using the reference image already facilitates this task. 
The corresponding cost curves, on the other hand, are characterised by the presence of wide and flat global minima as illustrated in Figure\,\ref{fig:CostCurves-SaddlePoint}. Thus, based on the cost curve the ambiguity arising from low textured or texture-less areas is clearly identifiable, allowing to improve the accuracy of the confidence estimation (c.f. Fig.\,\ref{fig:QualRes-Mv3-CVA}).

Finally, in Figure\,\ref{fig:EvalMv3} it can also be seen that most of CVA-Net's erroneous confidence estimates are close to depth discontinuities. It can be assumed that this is (at least partly) due to the fact that samples of depth discontinuities are underrepresented in the training set. First experiments with CVA-Net trained on synthetic data have shown that dense ground truth for the training images can help to overcome this limitation. Further investigations on that issue will be carried out in future work.
\section{Conclusion}
Inspired by the superior results of learned confidence measures on the one hand and confidence measures based on cost curve features on the other hand, in this paper, we propose to learn the estimation of confidence for dense stereo matching based on 3D cost volumes. 
We argue that such cost volumes contain additional information compared to disparity maps, which allows to estimate confidence more accurate.
To the best of our knowledge, this is the first time that complete cost volumes are used as input to a CNN in the context of this task.

With an extensive evaluation on three well-established datasets using three common stereo matching methods, we prove the superior performance of the proposed CVA-Net architecture compared to the state-of-the-art.
In the context of the evaluation, we furthermore discuss the weakness of the commonly used AUC computation approach of introducing a potentially significant discretisation error and propose a solution in form of an interval-based extension.
Finally, as already mentioned in Section\,\ref{sec:eval-kit}, there is space for further improvement: 
The sensitivity to depth discontinuities as well as to differences between curves from different stereo matching methods needs further investigation.
However, the results of the evaluation not only confirm the general validity of the proposed approach, but also demonstrate its superior accuracy especially in noisy regions.

\section*{Acknowledgements}
This work was supported by the German Research Foundation (DFG) as a part of the Research Training Group i.c.sens [GRK2159], the MOBILISE initiative of the Leibniz University Hannover and TU Braunschweig and by the NVIDIA Corporation with the donation of the Titan V GPU used for this research.

{\small
\bibliographystyle{ieee}

}

\end{document}